\documentclass{article} 
\usepackage{iclr2026_conference,times}

\usepackage{amsmath,amsfonts,bm}

\def\eqref#1{equation~\ref{#1}}

\def\1{\bm{1}}

\DeclareMathAlphabet{\mathsfit}{\encodingdefault}{\sfdefault}{m}{sl}
\SetMathAlphabet{\mathsfit}{bold}{\encodingdefault}{\sfdefault}{bx}{n}

\usepackage[T1]{fontenc}
\usepackage{inconsolata}

\usepackage{hyperref}
\usepackage{url}
\usepackage{booktabs}
\usepackage{graphicx}
\usepackage{multirow}
\usepackage{caption}
\usepackage{tcolorbox}
\tcbuselibrary{breakable}

\newtcolorbox{promptbox}[1][]{
  colback=gray!5, colframe=gray!40,
  fonttitle=\bfseries\small, title=#1,
  boxrule=0.4pt, left=3mm, right=3mm, top=1mm, bottom=1mm,
  fontupper=\small\ttfamily, breakable
}

\title{Can LLMs Perceive Time?\\An Empirical Investigation}

\author{Aniketh Garikaparthi \vspace{0.2em} \\ 
TCS Research \vspace{0.2em} \\
\texttt{aniketh.g@tcs.com}
}

\iclrfinalcopy 
\begin{document}

\maketitle

\begin{abstract}
Large language models cannot estimate how long their own tasks take. We investigate this limitation through four experiments across 68 tasks and four model families. Pre-task estimates overshoot actual duration by 4--7$\times$ ($p < 0.001$), with models predicting human-scale minutes for tasks completing in seconds. Relative ordering fares no better: on task pairs designed to expose heuristic reliance, models score at or below chance (GPT-5: 18\% on counter-intuitive pairs, $p = 0.033$), systematically failing when complexity labels mislead. Post-hoc recall is disconnected from reality---estimates diverge from actuals by an order of magnitude in either direction. These failures persist in multi-step agentic settings, with errors of 5--10$\times$. The models possess propositional knowledge about duration from training but lack experiential grounding in their own inference time, with practical implications for agent scheduling, planning and time-critical scenarios.
\end{abstract}

\section{Introduction}

As LLM agents increasingly tackle longer-horizon tasks \citep{kwa2025metr, wijk2025rebench, garikaparthi2026researchgymevaluatinglanguagemodel, liu2026klongtrainingllmagent}, they are also broadly deployed in settings that require planning, delegation, tool use, and coordination across multiple workers or sub-agents \citep{xie2024osworld, tran2025multiagentcollaboration, piskala2026maple}. In such systems, accurate temporal self-estimation becomes part of the control problem: a manager agent must decide whether to call a fast heuristic or a slower specialist, whether to parallelize or serialize branches, when to stop exploration and commit, and how to allocate limited test-time compute across subtasks \citep{Gray2023MeasuringTA, xu2025tpsbench, erdogan2025planandact, paglieri2025learningwhenplan, simhi2025managerbench}. These decisions become harder in nested settings, where one scheduler calls other schedulers and timing errors compound across levels. A system that cannot estimate how long its own actions take cannot schedule itself reliably.

This matters anywhere latency, deadlines, or scarce resources constrain behavior. In interactive computer-use settings, agents already operate under tool, budget, and execution constraints \citep{xie2024osworld, yao2025taubench, barres2025tau2bench}. In policy and control settings, punctuality is itself part of task success rather than a secondary concern \citep{jia2025timeawarepolicy, xu2024criticaltesting}. In time-critical domains such as medical triage, emergency response, and real-world resource allocation, misjudging duration can have direct operational consequences \citep{info:doi/10.2196/76326, sehgal2026realtimedeadlinesrevealtemporal, keding2021managerialoverreliance, sravanthi2023aiassistedresourceallocation}. If LLMs are to act as planners, orchestrators, or managers, they need at least coarse temporal calibration about their own behavior.

This question is related to, but distinct from, prior work on time in language models. Existing work has studied temporal fact recall, temporal reasoning over events, external time-series forecasting, temporal point processes, and spatio-temporal understanding in video \citep{ding2025understandtime, herel2024timeawarenessllms, jin2024timellm, chen2026temporalawarenesstpp, cheng2025vstar, song2025videommlu, zhao2025mmvu, liu2026bridgepredictinghumantask}. Those settings ask whether models can reason about time in the world. Our question is different: can a model estimate the duration of its own computation and actions? This is a problem of temporal self-estimation. It requires mapping internal processing and task demands to elapsed time, not merely recalling that one event happened before another or that a class of activities usually lasts a certain duration.


Humans develop such judgments through repeated interaction with the world. They act, wait, observe progress, and update expectations from feedback. That kind of calibration depends on continuous temporal grounding. A broad line of work in world models, predictive representation learning, embodied intelligence, and action-grounded systems argues that prediction and planning improve when a system learns from temporally extended interaction rather than text alone \citep{ha2018worldmodels, assran2023jepa, assran2025vjepa2, bruce2024genie, feng2025embodiedai, li2026whatmattersvla, lee2025molmoact, guo2026braininspired}. Standard LLM inference lacks this grounding. The model observes tokens, not elapsed time. It does not directly perceive wall-clock duration while generating, and it does not accumulate sensorimotor memories of how long similar acts took. The eventual duration of a response also depends on factors such as model size, hardware, batching, tool latency, and network effects, none of which are represented in the prompt. 

The picture becomes less stable as deployment increasingly decouples model identity from latency. Some providers now expose a single model under different speed modes.\footnote{\url{https://developers.openai.com/codex/speed/}}\footnote{\url{https://code.claude.com/docs/en/fast-mode}} Specialized inference hardware and newer sequence-model architectures make the mapping from semantic task difficulty to wall-clock duration even less direct. State-space models and diffusion language models change the underlying compute pattern of generation itself \citep{gu2023mamba, dao2024transformersssm, nie2025llada}. Runtime, therefore, is increasingly a property of the full deployment stack and architecture.

We investigate this limitation through four experiments across 68 tasks and four model families. We first study absolute calibration: before acting, can models estimate how long their own task execution will take? We then test relative judgment: even if absolute estimates are poor, can models at least identify which of two tasks will take longer? Next, we examine post-hoc recall: after completing a task, can a model report how long it actually took? Finally, we test whether the same limitation persists in multi-step agentic settings, where timing errors can propagate through planning, delegation, and resource allocation.

Our findings reveal a consistent pattern. Pre-task estimates overshoot actual duration by 4--7$\times$ (\(p < 0.001\)), often assigning human-scale minutes to tasks that finish in seconds. Relative judgments fare no better: on task pairs constructed to separate superficial complexity cues from true execution time, models perform at or below chance, indicating reliance on heuristics rather than temporal self-knowledge. Post-hoc estimates are similarly disconnected from reality, often differing from actual duration by an order of magnitude in either direction. The same failure persists in multi-step agentic settings, where errors remain in the 5--10$\times$ range. 

\section{The Problem: Missing Temporal Grounding}

LLMs cannot estimate their own task durations because they lack access to the necessary information. Table~\ref{tab:temporal-access} makes this concrete: models know human task durations from training data, and can receive timestamps via system prompts, but cannot access their own inference speed, elapsed generation time, or the mapping from tokens to seconds \cite{cheng2026llmagentstemporallyblind}.

This creates a fundamental asymmetry.
The model cannot estimate how long it will take \textit{itself} to write the same algorithm. Whether generating 200 tokens of code takes 4 seconds (A100 GPU), 20 seconds (consumer GPU with quantization), or 8 seconds (API with network latency) depends on deployment factors invisible during generation. Beyond self-estimation, agents must estimate external latencies such as tool execution, sub-agent calls, user response time, for which no grounding exists either.

\begin{table}[t]
\begin{minipage}[t]{0.48\textwidth}
\centering
\caption{Temporal information available to LLMs during generation.}
\label{tab:temporal-access}
\vspace{0.5em}
\begin{tabular}{lc}
\toprule
\textbf{Information} & \textbf{Available?} \\
\midrule
Human task durations & \checkmark \\
Current timestamp & \checkmark \\
\midrule
Own inference speed & $\times$ \\
Elapsed generation time & $\times$ \\
\bottomrule
\end{tabular}
\end{minipage}
\hfill
\begin{minipage}[t]{0.48\textwidth}
\centering
\caption{Absolute calibration across models (n=68 tasks per model). \footnotesize  **$p{<}.01$, ***$p{<}.001$}
\label{tab:exp1-results}
\vspace{0.5em}
\begin{tabular}{lcc}
\toprule
\textbf{Model} & \textbf{Med. Ratio} & \textbf{$r$} \\
\midrule
GPT-5 & 6.11$\times$ & 0.55*** \\
GPT-4o & 3.60$\times$ & 0.35** \\
\midrule
OLMo3-7B & 0.55$\times$ & $-$0.06 \\
Qwen3-8B & 0.78$\times$ & 0.18 \\
\bottomrule
\multicolumn{3}{l}{\footnotesize }
\end{tabular}
\end{minipage}
\end{table}

\section{Experiments}

We evaluate temporal self-estimation across 68 tasks spanning seven categories: code generation, debugging, summarization, reasoning, writing, creative, and question answering. Tasks range from trivial (``write hello world'') to very hard (``implement a regex engine''), with single-turn actual durations from 1--90 seconds depending on model and complexity. We test frontier API models (GPT-5, GPT-4o) and open-source models run locally on A100 GPUs (OLMo3-7B, Qwen3-8B). Details in Appendix~\ref{app:tasks}.

\subsection{Experiment 1: Absolute Calibration}

We first test whether pre-task duration estimates correlate with actual execution time. For each task, we prompt the model: ``How long will it take \textit{you} to complete this task? Give your estimate in seconds.'' We then execute the task and measure wall-clock duration from API request to response completion.

\begin{figure}[t]
\centering
\includegraphics[width=\textwidth]{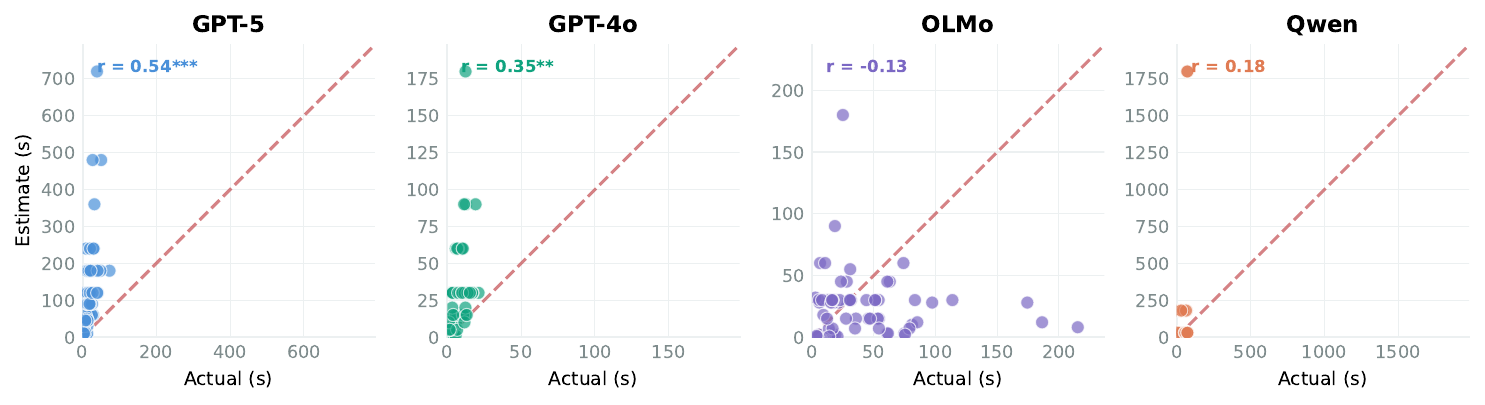}
\caption{Estimation calibration across models. Each point represents one task; dashed line indicates perfect calibration. Frontier models (GPT-5, GPT-4o) show weak positive correlation, while open models (OLMo, Qwen) cluster around arbitrary values with no relationship to actual duration.}
\label{fig:calibration}
\end{figure}

Table~\ref{tab:exp1-results} and Figure~\ref{fig:calibration} show the results. Frontier models exhibit moderate correlation---GPT-5 achieves $r = 0.55$, GPT-4o achieves $r = 0.35$---but with substantial bias: median estimates exceed actuals by 4--6$\times$. Models predict human-scale durations (tens of seconds to minutes) for tasks that complete in single-digit seconds. Open-source models show no significant correlation at all, with OLMo3-7B at $r = -0.06$ and Qwen3-8B at $r = 0.18$.

The pattern varies systematically with task complexity. Analyzing OLMo3-7B's predictions by complexity level reveals mild overestimation for simple tasks (trivial: 1.15$\times$, easy: 1.79$\times$) shifting to underestimation for complex tasks (hard: 0.69$\times$, very hard: 0.39$\times$), suggesting models anchor estimates on task descriptions rather than their own processing speeds. Additional breakdowns appear in Appendix~\ref{app:exp1}.

\subsection{Experiment 2: Relative Ordering}

Absolute calibration requires knowing both ranking and scale. Perhaps models can succeed at the simpler task of relative ordering: given two tasks, which takes longer?

Initial experiments with randomly selected pairs yielded near-perfect accuracy, but this proved trivially easy. Random pairs often had 2--4$\times$ duration differences, and simple heuristics (``code tasks take longer than summaries'') frequently succeeded. To create a meaningful test, we curated 26 ``hard pairs'' across three categories designed to defeat surface heuristics.

\textit{Near-identical pairs} (5 pairs) have less than 5\% duration difference---for example, \texttt{code\_fibonacci} versus \texttt{code\_linked\_list} at 5.63s and 5.69s respectively. These test whether models have any genuine signal beyond noise.

\textit{Counter-intuitive pairs} (11 pairs) are the critical diagnostic. Here, the ``harder'' complexity label corresponds to \textit{faster} actual completion---for example, \texttt{code\_lru\_cache} (labeled hard, 9.4s actual) versus \texttt{reason\_logic\_grid} (labeled medium, 10.1s actual). If models use complexity as a heuristic for duration, they will systematically fail on these pairs. Each pair was validated via empirical ground-truth runs (6 trials $\times$ position-swapped per pair), confirming that the higher-complexity task genuinely completes faster.

\textit{Cross-category pairs} (10 pairs) match different task types with similar durations, testing whether models use category as a proxy.

\begin{figure}[t]
\centering
\begin{minipage}[t]{0.38\textwidth}
\vspace{0pt}
\centering
\begin{tabular}{lccc}
\toprule
\textbf{Model} & \textbf{All} & \textbf{C-I} & \textbf{N-I} \\
\midrule
GPT-5 & 46\% & \textbf{18\%} & 60\% \\
GPT-4o & 58\% & 55\% & 80\% \\
OLMo3 & 46\% & 45\% & 60\% \\
Qwen3 & 54\% & 45\% & 60\% \\
\midrule
\textit{Chance} & 50\% & 50\% & 50\% \\
\bottomrule
\end{tabular}
\vspace{0.5em}
\\{\footnotesize C-I: counter-intuitive (n=11), N-I: near-identical}
\end{minipage}
\hfill
\begin{minipage}[t]{0.58\textwidth}
\vspace{0pt}
\centering
\includegraphics[width=\textwidth]{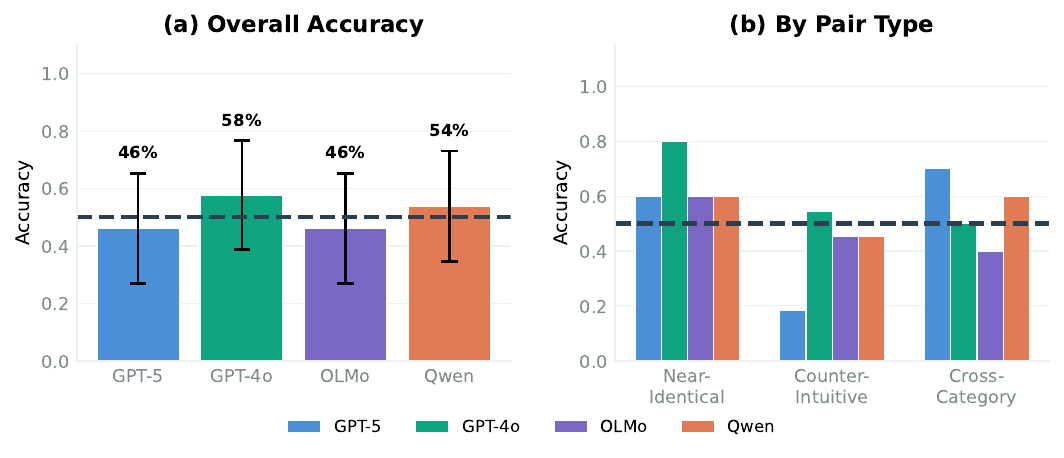}
\end{minipage}
\caption{Relative ordering accuracy on 26 hard pairs (left) and by pair type (right). Counter-intuitive pairs expose heuristic reliance---GPT-5 scores 18\% on 11 CI pairs ($p = 0.033$, one-sided binomial), significantly below chance. Overall accuracy ranges from 46--58\%, consistent with near-random performance on diagnostically hard pairs.}
\label{fig:ordering}
\end{figure}

Figure~\ref{fig:ordering} presents the results. Overall accuracy hovers near chance at 46--58\%. The counter-intuitive pairs provide the key finding: GPT-5 achieves only 18\% accuracy (2/11)---significantly below chance ($p = 0.033$, one-sided binomial). When complexity labels mislead, the model systematically chooses the wrong answer, demonstrating reliance on heuristics rather than genuine temporal self-knowledge. 

We also collected post-hoc ordering judgments after models completed both tasks. Post-hoc accuracy was not meaningfully better than prediction accuracy for any model, suggesting that models do not accumulate temporal information during generation. Full pair specifications in Appendix~\ref{app:hard-pairs}.

\subsection{Experiment 3: Post-Hoc Recall}

The previous experiment hints that models lack temporal memory during processing. We test this directly by asking models to estimate duration \textit{after} completing a task, without revealing any timing information.

Immediately after task completion, we prompt: ``You just completed this task. How long did it take you to generate that response? Estimate in seconds.'' The model has no external signal---no timestamps were provided, no duration was mentioned.

The results reveal a striking split between model families. Frontier models \textit{overestimate} post-hoc: GPT-4o claimed 42 seconds for tasks completing in 8 seconds (5.2$\times$), and GPT-5 estimated 32 seconds for 19-second tasks (1.7$\times$). Open-source models show variable patterns: OLMo3-7B estimated 30 seconds for tasks completing in 27 seconds (median 0.94$\times$), a near-match that is likely coincidental given $r = -0.06$ overall.

This pattern is informative. Frontier models appear to know that ``AI responses take time,'' producing human-plausible but uncalibrated durations. Open-source models produce estimates that may happen to coincide with actuals on average but show zero correlation across tasks ($r = -0.06$). Neither reflects actual processing time---the absence of correlation confirms zero temporal proprioception.

\subsection{Experiment 4: Agentic Tasks}

Real-world agents execute multi-step tasks with tool use. We tested whether temporal miscalibration extends to this setting using six agentic tasks: building a landing page, debugging a multi-file project, running a data analysis pipeline, creating a CLI tool, refactoring legacy code, and building a test suite. Each task used a ReAct agent with bash, Python, and text editor tools.

Pre-task estimates were 5--10$\times$ off, consistent with single-turn experiments. Post-hoc estimates were worse: GPT-4o claimed ``30 seconds'' for tasks that ran 10 minutes. The multi-step nature adds additional uncertainty---models cannot predict tool latency, retry loops, or debugging time. Task success or failure did not affect calibration; even successful completions showed large estimation errors. Full per-task breakdowns appear in Appendix~\ref{app:agentic}.


\section{Related Work}

\paragraph{Long-horizon and multi-step agents.}
LLM agents have moved well beyond single-turn text generation. Recent work studies increasingly long-horizon task execution in interactive environments, software engineering, real computer use, and broader agent benchmarks \citep{kwa2025metr, manakina2025delayofgratification, jiang2026agentlabbenchmarkingllmagents, wijk2025rebench, xi2026agentgymrl, garikaparthi2026researchgymevaluatinglanguagemodel, liu2026klongtrainingllmagent}. Parallel work studies coordination and specialization in multi-agent and hierarchical systems \citep{yao2025taubench, barres2025tau2bench, tran2025multiagentcollaboration, lazaridou2017multiagentcooperation, lazaridou2020multiagentcommunication, ruan2026aorchestra, estornell2025trainleader}. As these systems become deeper and more nested, scheduling, routing, and test-time compute allocation become part of the core control problem rather than an implementation detail \citep{Gray2023MeasuringTA, erdogan2025planandact, paglieri2025learningwhenplan}. Our work focuses on a basic capability required in such settings: estimating how long the agent's own actions take.

\paragraph{Time in language models.}
A growing literature studies time-related capabilities in language models, but mostly in senses other than self-duration. Existing work examines temporal fact recall, event ordering, temporal validity across time, temporal point processes, and external time-series forecasting \citep{ding2025understandtime, herel2024timeawarenessllms, goel2025chronoceptinstillingsensetime, chen2026temporalawarenesstpp, jin2024timellm}. In multimodal settings, recent benchmarks study spatio-temporal reasoning in video and report related failures of temporal sensitivity \citep{cheng2025vstar, song2025videommlu, zhao2025mmvu, upadhyay2025timeblindnessvideolanguagemodels}. Concurrent work also studies temporal blindness in agent tool-use decisions, asking whether agents act as if time has passed 
\citep{cheng2026llmagentstemporallyblind}. 

\paragraph{Grounding, embodiment, and world models.}
Why might self-duration be hard? A natural hypothesis is that it depends on forms of temporal grounding that text-only pretraining does not provide. Work on world models, predictive representation learning, embodied intelligence, and vision-language-action systems argues that effective prediction and planning benefit from temporally extended interaction, state tracking, and feedback from the consequences of actions \citep{ha2018worldmodels, assran2023jepa, assran2025vjepa2, bruce2024genie, feng2025embodiedai, li2026whatmattersvla, lee2025molmoact, guo2026braininspired, jia2025timeawarepolicy}. By contrast, standard LLM inference is typically given text, not direct access to elapsed time. And timing is often represented only indirectly through step counts, token counts, timeout wrappers, or prompt-level timestamps. These are useful control signals, but they are ad hoc substitutes for continuous temporal perception.

\paragraph{Introspection and self-knowledge.}
Our work also relates to emerging work on LLM introspection. Recent studies suggest that models can, under suitable training, access limited information about their own behavioral tendencies or some injected internal states, but that these abilities remain narrow and brittle \citep{binder2024introspection, lindsey2026introspection}. 

\section{Conclusion}

Large language models cannot accurately estimate their own task durations. Frontier models show weak correlation with actuals ($r = 0.35$--$0.55$) while open models show none. Counter-intuitive task pairs reveal that models rely on complexity heuristics---GPT-5 scores 18\%, below chance, when harder-labeled tasks are actually faster. Post-hoc recall is disconnected from reality, with models believing they completed tasks much slower or, in some cases, much faster. These limitations persist in multi-step agentic settings. The architectural limitation requires deeper solutions beyond scaffolding and introducing timestamps through external infrastructure. Future work should explore training with explicit timing signals and architectures that better retain and use temporally grounded state, especially for real-world deadline-sensitive deployments.

\bibliography{references}
\bibliographystyle{iclr2026_conference}

\newpage
\appendix

\section{Task Suite}
\label{app:tasks}

The 68 tasks span seven categories designed to cover realistic LLM use cases. Code generation includes 12 tasks from hello world to regex engine implementation. Debugging covers 8 tasks including off-by-one errors, memory leaks, and async deadlocks. Summarization spans 10 tasks from short paragraphs to technical documents. Reasoning includes 12 tasks from arithmetic word problems to algorithm analysis. Writing covers 10 tasks from tweets to essays. Creative includes 8 tasks from haikus to worldbuilding. Question answering covers 8 tasks from factual queries to paper summaries.

Complexity levels correspond roughly to actual duration ranges for frontier API models: trivial (1--3s), easy (3--8s), medium (8--20s), hard (20--40s), and very hard (40--90s). Open-source models running locally may exhibit substantially longer durations, particularly for complex tasks. We tested frontier API models (GPT-5, GPT-4o from OpenAI) and open-source models run locally (OLMo3-7B, Qwen3-8B via Hugging Face Transformers on A100 GPUs).

\section{Experiment 1: Full Results}
\label{app:exp1}

\begin{table}[h]
\begin{minipage}[t]{0.55\textwidth}
\centering
\caption{Full absolute calibration results.}
\vspace{0.5em}
\begin{tabular}{lccc}
\toprule
\textbf{Model} & \textbf{Mean} & \textbf{Median} & \textbf{$r$} \\
\midrule
GPT-5 & 7.03$\times$ & 6.11$\times$ & 0.55*** \\
GPT-4o & 4.41$\times$ & 3.60$\times$ & 0.35** \\
OLMo3-7B & 1.48$\times$ & 0.55$\times$ & $-$0.13 \\
Qwen3-8B & 2.66$\times$ & 0.78$\times$ & 0.03 \\
Qwen3 (think) & 1.23$\times$ & 0.59$\times$ & 0.44** \\
\bottomrule
\end{tabular}
\end{minipage}
\hfill
\begin{minipage}[t]{0.42\textwidth}
\centering
\caption{Log-scale correlations.}
\vspace{0.5em}
\begin{tabular}{lc}
\toprule
\textbf{Model} & \textbf{$r$ (log)} \\
\midrule
GPT-5 & 0.65*** \\
GPT-4o & 0.45*** \\
OLMo3-7B & $-$0.03 \\
Qwen3-8B & 0.15 \\
\bottomrule
\end{tabular}
\end{minipage}
\end{table}

Calibration varies systematically with task complexity. OLMo3-7B overestimates simple tasks (trivial: 1.15$\times$, easy: 1.79$\times$) and underestimates complex ones (medium: 0.87$\times$, hard: 0.69$\times$, very hard: 0.39$\times$). This pattern suggests anchoring on task descriptions rather than actual processing characteristics.

Qwen3-8B with thinking mode enabled \textit{underestimates} duration (median 0.59$\times$) because it does not account for reasoning overhead. Without thinking mode, the same model \textit{overestimates} (median 0.78$\times$). The model estimates based on task complexity, not its actual processing.

\section{Hard Pairs Methodology}
\label{app:hard-pairs}

Random task pairs yielded near-100\% accuracy because large duration gaps (2--4$\times$) made correct ordering trivial. We curated 26 hard pairs: 5 near-identical, 11 counter-intuitive, 10 cross-category.

\begin{table}[h]
\centering
\caption*{\textbf{Counter-Intuitive Pairs} (harder label, faster actual; validated 6 runs $\times$ position-swap)}
\vspace{0.3em}
\begin{tabular}{llcc}
\toprule
\textbf{Task A} & \textbf{Task B} & \textbf{Complexity} & \textbf{Actual (A/B)} \\
\midrule
lru\_cache & logic\_grid & hard / med & 9.4s / 10.1s \\
book\_chapter & creative\_rewrite & hard / med & 3.85s / 3.72s$^\dagger$ \\
regex\_engine & qa\_historical & v.hard / med & 12.6s / 13.1s \\
async\_deadlock & story\_medium & hard / med & 8.3s / 8.4s \\
\midrule
game\_theory & write\_tech\_doc & hard / med & 19.3s / 26.7s \\
medical\_report & story\_short & hard / easy & 15.1s / 22.2s \\
optimization & write\_user\_story & hard / med & 23.3s / 26.0s \\
legal\_doc & analogy & med / easy & 9.2s / 19.7s \\
research\_paper & age\_problem & med / easy & 10.2s / 20.5s \\
fizzbuzz & swap\_variables & easy / trivial & 4.7s / 20.3s \\
news\_article & email\_prof. & easy / trivial & 5.8s / 9.7s \\
\bottomrule
\multicolumn{4}{l}{\footnotesize $^\dagger$Near-equal; direction confirmed via majority vote across 6 position-swapped runs.}
\end{tabular}
\end{table}

All pairs were validated via ground truth establishment: multiple runs per pair with A/B position swapping to cancel prompt-position bias. Ground truth is the majority winner. Counter-intuitive pairs share a common pattern: the harder-labeled task has more constrained, structured output (extracting, solving a defined problem) while the easier-labeled task requires open-ended generation.

\section{Ablations}
\label{app:ablations}

\subsection{Reasoning Effort}

GPT-5 supports configurable reasoning effort levels. We tested whether increased reasoning improves calibration by running all 68 tasks at four effort levels.

\begin{table}[h]
\centering
\caption{GPT-5 calibration by reasoning effort level.}
\begin{tabular}{lcc}
\toprule
\textbf{Effort} & \textbf{Mean Ratio} & \textbf{Median Ratio} \\
\midrule
Minimal & 10.01$\times$ & 7.78$\times$ \\
Low & 10.60$\times$ & 8.87$\times$ \\
Medium & 6.79$\times$ & 6.05$\times$ \\
High & 4.94$\times$ & 3.78$\times$ \\
\bottomrule
\end{tabular}
\end{table}

\begin{figure}[h]
\centering
\includegraphics[width=0.8\textwidth]{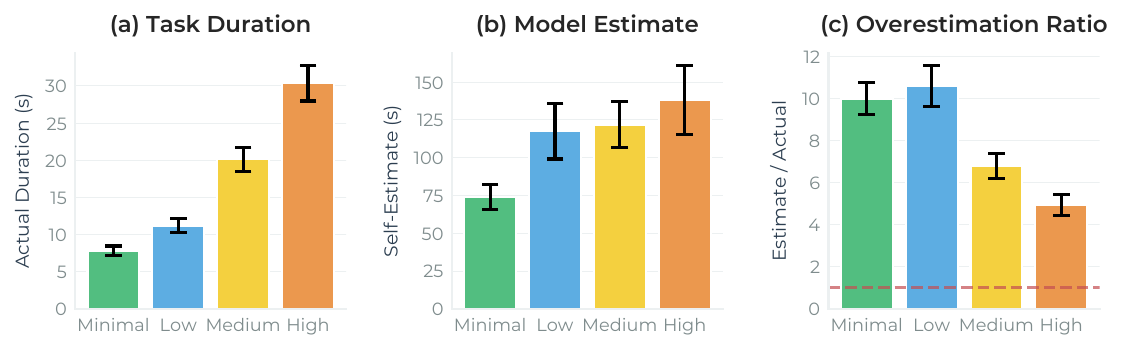}
\caption{GPT-5 calibration by reasoning effort level. Higher effort reduces overestimation ratio (left) as actual duration increases to match human-scale estimates. Correlation improves slightly (right) but remains driven by task complexity, not self-awareness.}
\label{fig:reasoning-effort}
\end{figure}

Higher reasoning effort improves calibration because more tokens require more time, incidentally approaching human-scale durations. This is not genuine self-awareness---the model cannot predict how much reasoning it will perform or map reasoning to wall-clock time. This finding is consistent with test-time compute scaling research \citep{snell2024scaling}: while additional reasoning improves task performance, models remain unaware of the computational cost they incur. The same limitation applies to reasoning models trained via reinforcement learning \citep{deepseek2025r1}.

\subsection{In-Context Calibration Feedback}

We tested whether explicit calibration feedback could improve estimates. The design: (1) baseline phase---estimate and execute 5 tasks spanning trivial to very hard; (2) feedback phase---inform the model of its error on one task (e.g., ``You estimated 120s, actual was 28.8s---you overestimated by 4.2$\times$''); (3) test phase---estimate and execute 5 \textit{new} tasks with feedback in context.

On GPT-5 with feedback from a medium-complexity task, baseline estimates averaged 13.2$\times$ overestimation while test-phase estimates averaged 2.3$\times$. However, the test tasks differed from baseline tasks, confounding direct comparison. More telling: even with explicit feedback in context, GPT-5 still produced 2--4$\times$ errors on individual tasks. The model cannot reliably apply ``I am $N\times$ off'' as a correction factor, suggesting the limitation is not purely epistemic.

\subsection{Thinking Mode (Qwen3-8B)}

Qwen3-8B supports an optional thinking mode that generates explicit reasoning before responding. We ran all 68 tasks with and without thinking on identical local hardware (A100 GPU).

Without thinking: median ratio 0.78$\times$, $r = 0.03$ (not significant). With thinking: median ratio 0.59$\times$, $r = 0.44$ ($p < 0.01$). Thinking mode \textit{improves} correlation---the model better ranks tasks by relative duration---but \textit{worsens} absolute calibration, underestimating by 1.7$\times$ instead of 1.3$\times$.

The model does not account for its own reasoning overhead. Thinking mode adds 3--5$\times$ more output tokens (and proportionally more time), yet estimates remain anchored on task complexity rather than actual processing. This confirms the architectural limitation: models cannot observe or predict their own computational overhead.

\begin{figure}[h]
\centering
\includegraphics[width=0.8\textwidth]{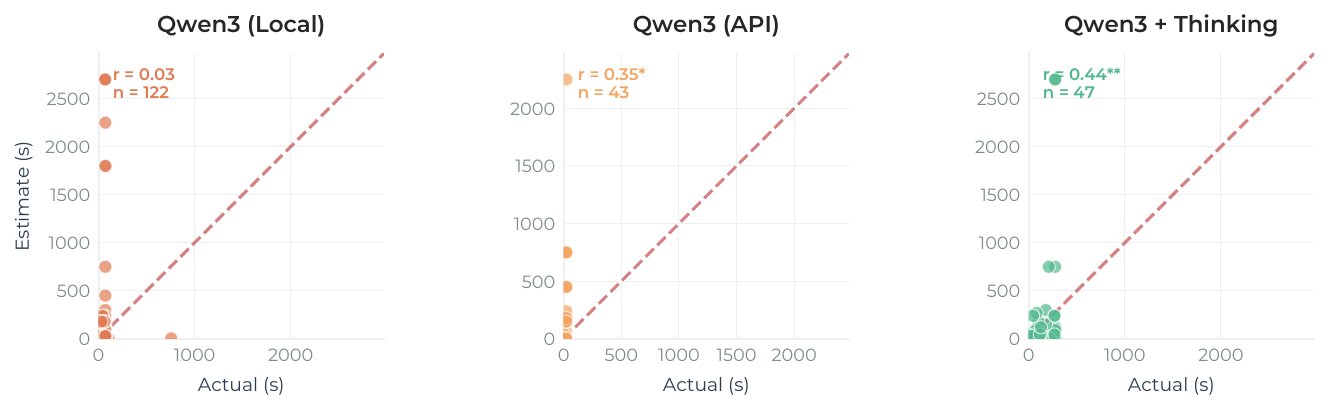}
\caption{Qwen3-8B calibration with and without thinking mode. Thinking improves correlation ($r = 0.44$ vs $r = 0.18$) but increases underestimation as the model fails to account for reasoning overhead.}
\label{fig:qwen-ablation}
\end{figure}

\section{Agentic Task Details}
\label{app:agentic}

Six multi-step tasks using ReAct agent with bash, Python, and text editor: build landing page, debug multi-file project, data analysis pipeline, build CLI tool, refactor legacy code, build test suite.

\begin{table}[h]
\begin{minipage}[t]{0.48\textwidth}
\centering
\caption{GPT-5 agentic results (100\% success).}
\vspace{0.3em}
\begin{tabular}{lccc}
\toprule
\textbf{Task} & \textbf{Act.} & \textbf{Pre} & \textbf{Post} \\
\midrule
Landing & 53s & 6.8$\times$ & 13$\times$ \\
Debug & 44s & 9.5$\times$ & 6.8$\times$ \\
Data & 41s & 5.9$\times$ & 4.4$\times$ \\
CLI & 75s & 8.0$\times$ & 0.4$\times$ \\
Refactor & 67s & 7.1$\times$ & 28$\times$ \\
Tests & 57s & 26$\times$ & 4.8$\times$ \\
\bottomrule
\end{tabular}
\end{minipage}
\hfill
\begin{minipage}[t]{0.48\textwidth}
\centering
\caption{GPT-4o agentic results (33\% success).}
\vspace{0.3em}
\begin{tabular}{lccc}
\toprule
\textbf{Task} & \textbf{Act.} & \textbf{Pre} & \textbf{Post} \\
\midrule
Landing & 8s & 11$\times$ & 1.2$\times$ \\
Debug & 12s & 15$\times$ & 2.5$\times$ \\
Data & 616s & 0.4$\times$ & 0.05$\times$ \\
CLI & 615s & 0.8$\times$ & 0.05$\times$ \\
Refactor & 609s & 0.5$\times$ & 0.01$\times$ \\
Tests & 611s & 2.0$\times$ & 0.02$\times$ \\
\bottomrule
\end{tabular}
\end{minipage}
\end{table}

Pre-estimates are 5--10$\times$ off, consistent with single-turn experiments. Post-hoc estimates are completely disconnected: GPT-4o claimed ``30 seconds'' for tasks that ran 10 minutes. Task success or failure does not affect calibration---even successful completions show large errors. Multi-step execution adds uncertainty from unpredictable tool latency, retries, and debugging loops.

\begin{figure}[h]
\centering
\includegraphics[width=0.9\textwidth]{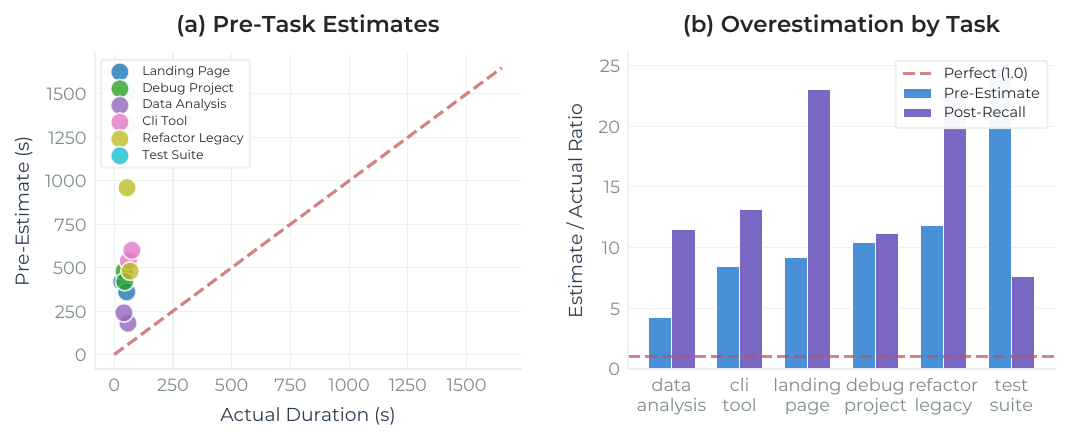}
\caption{Agentic task estimation errors. Pre-task estimates (left) overshoot by 5--10$\times$. Post-hoc estimates (right) show even larger disconnection from actual duration, with GPT-4o's failed tasks producing extreme underestimates.}
\label{fig:agentic}
\end{figure}

\section{Implementation Details}
\label{app:implementation}

\subsection*{Timing}
Wall-clock duration is measured from API request initiation to response completion, including network latency, reflecting the duration an agent system would observe. For local models, timing runs from generation start to completion. One warmup request precedes each session to reduce cold-start variance. Duration estimates are extracted via regex supporting explicit formats (``30 seconds''), ranges (``1--2 minutes,'' midpoint taken), and vague expressions (``a few seconds'' $\to$ 3s).

\subsection*{Prompts}

\begin{promptbox}[Pre-task estimation (Exp 1--2)]
Task: \{description\}\\
How long will it take you to complete this? Give a specific estimate in seconds.\\
Reply with just the number and unit, e.g. "30 seconds" or "2 minutes".
\end{promptbox}

\begin{promptbox}[Post-hoc recall (Exp 3--4)]
You just completed this task: \{description\}\\
Your response: \{output\}...\\
How long did it take you to generate that response?\\
Give your best estimate in seconds. Reply with just the number and unit.
\end{promptbox}

\begin{promptbox}[Relative ordering (Exp 2)]
I'm going to ask you to complete two tasks. Which one will take YOU (the AI) longer to complete?\\
Task A: \{description\_A\}\\
Task B: \{description\_B\}\\
Reply with just "A" or "B".
\end{promptbox}

\subsection*{Model configuration}
All API models use provider default temperatures (GPT-5/4o: 1.0). Local models (OLMo3-7B, Qwen3-8B) run on A100 GPUs with \texttt{max\_tokens=2048}, \texttt{repetition\_penalty=1.1}, and explicit \texttt{eos\_token\_id} to prevent runaway generation. Qwen3-8B requires \texttt{temperature} $\geq 0.6$ per model card. Experiment settings: 5 runs per task, 120s timeout, 3 retries with exponential backoff (factor 2.0). Full config in \texttt{config.yaml}.

\section{Extended Discussion}
\label{app:discussion}

The gap between knowing \textit{about} time and knowing one's \textit{own} time proves consistent across experiments. Models possess duration knowledge from training---they can reasonably estimate how long tasks take humans---but this knowledge does not transfer to self-estimation. The mapping from ``task complexity'' to ``my inference time'' simply does not exist in training data and cannot be learned without access to timing information during training or inference.

The counter-intuitive pairs provide the clearest evidence. If models had genuine temporal self-awareness, they would not systematically fail when complexity labels mislead. GPT-5's 18\% accuracy on 11 diagnostic CI pairs (2/11, $p = 0.033$)---significantly below chance---demonstrates that even the best-calibrated model relies on heuristics. The moderate absolute correlation ($r = 0.55$) likely reflects learned relationships between task descriptions and response lengths, not temporal perception.

The frontier-versus-open gap deserves attention. GPT-5 and GPT-4o show weak but significant correlation; OLMo3-7B and Qwen3-8B show none. Larger models may have learned some calibration signal from the relationship between task complexity and typical response length. However, this signal remains insufficient for practical scheduling---even GPT-5 overestimates by 4--6$\times$ and fails significantly below chance on counter-intuitive pairs.

\paragraph{Practical implications.} For agent system designers: do not rely on model self-estimation for scheduling. Effective approaches include external timing infrastructure that tracks and reports elapsed time, historical logging of actual durations for task types to enable lookup-based estimation, explicit time budgets communicated to the model, and timeout mechanisms at the system level rather than relying on model self-regulation.

\paragraph{Limitations.} We test specific models on 68 English-language tasks; different domains or languages may show different patterns. We focus on correlation as our primary metric; future work could examine whether architectural changes (timing tokens, compute-aware training) could provide the missing grounding.

\end{document}